\def\year{2018}\relax
\documentclass[letterpaper]{article} 
\usepackage{aaai18}  
\usepackage{times}  
\usepackage{helvet}  
\usepackage{courier}  
\usepackage{url}  
\usepackage{graphicx}  

\usepackage{epsfig} 
\usepackage{amsmath}
\usepackage{amssymb}
\usepackage{subfigure}
\usepackage{amsopn}
\usepackage{mathrsfs}
\usepackage{xcolor}
\usepackage{array}
\usepackage{algorithm}
\usepackage{algorithmic}

\begin{document}
%
\title{Learning to Guide Decoding for Image Captioning}
\author{Wenhao Jiang$^{1}$\quad Lin Ma$^{1}$\quad Xinpeng Chen$^{2}$\quad Hanwang Zhang$^{3}$\quad Wei Liu$^{1}$ \\ 
$^{1}$Tencent AI Lab, 
$^{2}$Wuhan University, 
$^{3}$Nanyang Technological University\\
\{cswhjiang, forest.linma\}@gmail.com, xinpeng\_chen@whu.edu.cn, \\
hanwangzhang@gmail.com, wliu@ee.columbia.edu 
}
\maketitle
\begin{abstract}
Recently, much advance has been made in image captioning, and an encoder-decoder framework has achieved outstanding performance for this task. In this paper, we propose an extension of the encoder-decoder framework by adding a component called guiding network. The guiding network models the attribute properties of input images, and its output is leveraged to compose the input of the decoder at each time step. The guiding network can be plugged into the current encoder-decoder framework and trained in an end-to-end manner. Hence, the guiding vector can be adaptively learned according to the signal from the decoder, making itself to embed information from both image and language. Additionally, discriminative supervision can be employed to further improve the quality of guidance. The advantages of our proposed approach are verified by experiments carried out on the MS COCO dataset. 
\end{abstract}

\section{Introduction}
Recently, much progress has been made in the high-level vision problems, such as image captioning \cite{vinyals2015show,xu2015show,fang2015captions}. Generating descriptions of images automatically is very useful in practice, for example, it can help visually impaired people understand image contents and improve image retrieval quality by discovering salient contents. 

Even though a great success has been achieved in object recognition, 
describing the contents of images automatically is still a very challenging task and more difficult than visual classification.  Image captioning models need to have a thorough understanding of a certain image, and capture the complicated relationships among objects and find salient ones. Moreover, they have to understand the interactions between images and languages and translate the image representations into language representations. Finally, the sentences  generated to express the captured information need to be natural.

Inspired by the advance in machine translation, 
the encoder-decoder framework was proposed to describe image contents automatically. The encoder is usually a convolutional neural network (CNN), while decoder is a recurrent neural network (RNN). The encoder represents the image with a vector, which captures objects and semantic information, while decoder generates natural sentences by consuming the image representation. However, this vector is unlikely to capture all necessary structural information needed for generating the description in the subsequent decoding phase. Moreover, decoder steps need to generate sentences that both describe the image contents and fit the trained language model. The decoder needs to balance the two factors to avoid that either one dominates the generating process. We noticed that sometimes the generated sentence seems to be too common and only weakly related to the input image. 

One direction to overcome the above limitations of the encoder-decoder framework is to introduce attention mechanisms, which steer the captioning model to focus on the important information for generating image descriptions \cite{xu2015show}. Models with attention have achieved promising performance on the image captioning tasks and some variants \cite{lchen_attention2016,yang2016review} have also been proposed by providing better annotation vectors. Useful external information related to input images has also been introduced into the attention model to improve the performance, including \cite{you2016image,jmun_aaai2017}.


Another direction to address the limitations of the encoder-decoder framework is to inject useful information into the decoder at each time step. Because the decoder can access the information related images at each time step, balancing between describing the input image and fitting the language model becomes easier. Some works  have followed this direction and introduced  predictions of word occurrences in captions \cite{yao2016boosting}, semantic vector \cite{jia_iccv2015,zhou2016image} into the decoder step, and promising performance have been presented. However, none of these models could learn the injected information for the decoder automatically. Hence, it could not be adaptively adjusted according to the decoder.

In this paper, we propose a novel model that can learn an extra guiding vector for the sentence generation progress automatically. The guiding vector is modeled with a guiding neural network, which is plugged into the decoder and accepts error signals from the decoder at each time step. Hence, it can be trained in a complete end-to-end way. Moreover, supervision from captions can be employed to perform discriminative training to learn a better guidance.  

\section{Related Works}\label{related_work}
The related works about image captioning can be divided into three categories. We give a brief review in this section.

\paragraph{Template-based methods.}
In the early stage of the research area of visual to language, template based methods were proposed to describe image contents automatically 
\cite{farhadi2010every,kulkarni2013babytalk}. 
This kind of methods detect attributes, scenes and objects first and fill in a sentence template, usually a triplet. Then sentences are generated with these templates.  Farhadi $et~al.$ \cite{farhadi2010every} inferred a triplet of scene elements which is converted to text using templates. Kulkarni $et~al.$ \cite{kulkarni2013babytalk} adopted a conditional random field (CRF) to jointly reason across objects, attributes, and prepositions before filling the templates. \cite{kuznetsova2012collective,mitchell2012midge} use more powerful language templates, such as a syntactically well-formed tree, and add descriptive information from the output of attribute detection.

\paragraph{Retrieval-based methods.}  Retrieval-based methods \cite{kuznetsova2012collective,mason2014nonparametric} first retrieve visually similar images, and then transfer captions of those images to the target image. The advantage of these methods is that the generated captions are more natural than the generations by template-based methods. However, because they directly rely on retrieval results among training data, there is little flexibility for them to add or remove words based on the content of an image.

\paragraph{Neural network-based methods.}
Recently, inspired by advance in machine translation, the encoder-decoder framework 
has also been applied to image caption generations \cite{vinyals2015show,xu2015show,yang2016review}. In this framework, a convolutional neural network (CNN)  trained for an image classification task is used as the image encoder, with the last fully-connected layer as the input to a recurrent neural network (RNN) decoder, which generates each word step by step until the end of the sentence \cite{vinyals2015show}. In \cite{xu2015show}, an attention mechanism was introduced into this framework. The attention mechanism could help the model focus on the subregions that are important for the current time step. 
 In \cite{yang2016review}, review steps were proposed to be inserted in the middle of the encoder and decoder, as an extension of \cite{xu2015show}. The review steps can learn the annotation vectors and initial states for the decoder steps, and the learned annotations and initial states are better than those generated by the encoder. 

Except those extensions to attention mechanisms, several models have been proposed to improve the image captioning performance by introducing more useful information into the encoder-decoder framework. In \cite{wu2016value}, the word occurrence prediction was treated as a multi-label classification problem and a region-based multi-label classification framework was proposed to extract visually semantic information. This prediction is then used to initialize a memory cell of a long short-term memory (LSTM) model. Yao $et~al.$ discussed different approaches to incorporate word occurrence predictions into the decoder step \cite{yao2016boosting}. 

Recently, policy-gradient methods for reinforcement learning were combined with the encoder-decoder framework. In \cite{rennie2016self}, the cross entropy loss was replaced with non-differentiable test metrics for image captioning tasks, e.g., CIDEr, and the system was trained with the REINFORCE algorithm~\cite{williams1992simple}. Since it optimized the testing metrics directly, significant improvements were achieved. This method can be used to improve the performance of all encoder-decoder models.

\section{Background}
Our method is an extension to the encoder-decoder framework. We provide a short review of encoder-decoder framework in this section. 

\paragraph{Encoder.} Under the encoder-decoder framework for image captioning, a CNN is usually employed as encoder to extract global representations and subregion representations. Global representations are usually the outputs of full connecting layers and subregion representations are usually the outputs of convolutional layers. Many different CNN can be used, e.g., VGG \cite{Simonyan14c}, Inception V3 \cite{szegedy2016rethinking}, ResNet \cite{he2016deep}. In this paper, we use Inception V3 as encoder. And the extracted global representation and subregion representations are denoted as $\mathbf{a}_0$ and $\mathcal{A} = \{\mathbf{a}_1, \dots, \mathbf{a}_k\}$ respectively, where $k$ is the number of kernels. In this paper, $\mathbf{a}_0$ is the output of last full-connected layer and $\mathcal{A}$ are the outputs of the last convolutional layer (before pooling).

\paragraph{Decoder.} Given the image representations, a decoder is employed to translate the image into natural sentences. A decoder is a recurrent neural network (RNN), and gated recurrent unit (GRU) \cite{chung2014empirical} 
or long short-term memory \cite{hochreiter1997long} (LSTM) can be utilized. In this paper, we use LSTM as the basic unit of the decoder. To overcome the limitations of LSTM, attention mechanisms were introduced to help the decoders focus on important parts at each time step. Attentive models can easily outperform non-attentive models. Hence, we only consider the decoders with attention mechanisms in this paper.

Recall that a LSTM with attention mechanism is a function that outputs the results based on the current hidden state, current input, and context vector. Context vector is the weighted sum of annotation vectors, and the weights are determined by a attention model. We adopt the same LSTM used in \cite{xu2015show} and express the LSTM unit with attention as follows:
\begin{align}
\left( \begin{array}{c}
\mathbf{i}_{t}  \\
\mathbf{f}_{t} \\
\mathbf{o}_{t} \\
\mathbf{g}_{t} \\
\end{array} \right) &=
\left( \begin{array}{c}
\sigma  \\
\sigma \\
\sigma \\
\textrm{tanh} \\
\end{array} \right)
T
\left( \begin{array}{c}
\mathbf{x}_t  \\
\mathbf{h}_{t-1} \\
\hat{\mathbf{z}}_{t-1} \\
\end{array} \right), \\
\mathbf{c}_t &= \mathbf{f}_t \odot \mathbf{c}_{t-1} + \mathbf{i}_{t}\odot \mathbf{g}_t, \\
\mathbf{h}_t &= \mathbf{o}_t \odot \textrm{tanh}(\mathbf{c}_t),
\end{align}
where $\mathbf{i}_t$, $\mathbf{f}_t$, $\mathbf{c}_t$, $\mathbf{o}_t$ and $\mathbf{h}_t$ are input gate, forget gate, memory cell, output gate and hidden state of the LSTM, respectively. Here, $T$ is a linear transformation operator,  $\mathbf{x}_t$ is the input signal at the $t$-th time step and $\hat{\mathbf{z}}_t $ is the context vector which is the output of attention model $f_{\textrm{att}}(\mathcal{A}, \mathbf{h}_{t-1})$.


The aim of image captioning is to generate a caption $\mathcal{C}$ consisting of $N$ words $(y_1,y_2,\cdots, y_N)$ given a certain image $\mathcal{I}$. The objective adopted is usually to minimize a negative log-likelihood:
\begin{align}\label{obj_lm}
\mathcal{L} = -\log\ p(\mathcal{C}|\mathcal{I}) = -\sum_{t=0}^{N-1}  \log p(y_{t+1}|y_{t}),	
\end{align}
where
$p(y_{t+1}|y_{t}) = \textrm{Softmax}(\mathbf {W}\mathbf{h}_t)$  and
$\mathbf{h}_t = \textrm{LSTM}(\mathbf{E}\mathbf{y}_{t}, \mathbf{h}_{t-1},f_{\textrm{att}}(\mathcal{A}, \mathbf{h}_{t-1}))$.
Here, $\mathbf{W}$ is a matrix for linear transformation, $y_0$ is the sign for the start of sentences, and $\mathbf{E}\mathbf{y}_{t}$ denotes the distributed representation of the word $\mathbf{y}_{t}$, in which $\mathbf{y}_{t}$ is the one-hot representation for the word $y_{t}$ and $\mathbf{E}$ is the word embedding matrix.  Let $f_{\textrm{att}}(\mathcal{A}, \mathbf{h}_{t-1})$ denote the attention model which usually attends to different regions of image $\mathcal{I}$ \cite{xu2015show} during the step-by-step word generation process. The objective of such an attention model is to guide the model to focus on the important subregions at the current time step.

\section{The Proposed Methods}\label{model}

\begin{figure*}
 \centering
  \includegraphics[width=0.8\linewidth]{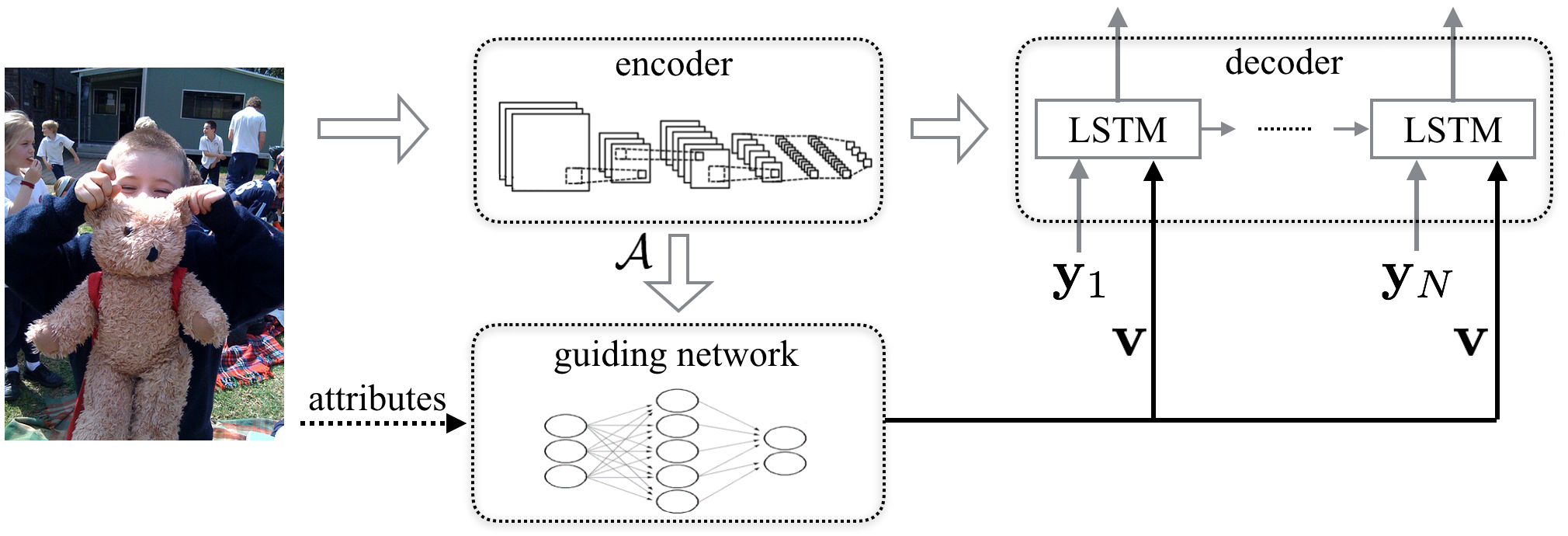}
  \caption{An illustration of learning to guide decoding for LSTM with attention. A guiding network is used to learn the guiding vector $\mathbf{v}$, which is used to compose the input of decoder with current word. The inputs of guiding network are the attribute features extracted from input image and the annotation vectors $\mathcal{A}$ provided by encoder.}
  \label{illustration}
\end{figure*}

In this section, we will present our model. For the existing encoder-decoder models, the input of decoder at each time step is usually the word embedding of previous word. Here, we consider to learn an informative guiding vector which can be used to compose the input for the decoder at each time step. The framework is shown in Fig.~\ref{illustration}. Here, we use a guiding network to model the guiding vector $\mathbf{v}$. Its inputs are annotations vectors from encoder and attribute features extracted from input image. The guiding network is plugged as a component into the current encoder-decoder framework  and can be trained in an end-to-end way. Details are described in the following subsections.

\subsection{Leanring To Guide Decoding}
In the decoder step, image representations are translated into sentences. The generated sentences need to not only describe the image contents, but also fit the trained language model. Auxiliary guiding information about images or captions injected in the decoder step could help to generate better sentences \cite{jia_iccv2015}. Here, we propose to learn guiding information from image representations or other extra inputs automatically.

\subsubsection{Guiding Network}\label{guiding_network}
In the decoder step of the conventional encoder-decoder framework, the input signal $\mathbf{x}_t$ is usually just $\mathbf{E}\mathbf{y}_{t}$, however if we can inject guiding information about the input image, we can get better performance. For example, we can simply use 
\begin{align}\label{input_of_decoder}
\mathbf{x}_t = \mathbf{E}\mathbf{y}_t + \mathbf{W}_v \textbf{v},
\end{align}
where guiding vector $\textbf{v}$ is a vector that contains information of the whole image and does not change in the process of generation, which can be seen a kind of hint of captions. For example, $\textbf{v}$  can be $\mathbf{a}_0$ \cite{jia_iccv2015} or predictions of word occurrences in captions \cite{yao2016boosting}. Here, we propose to learn the vector by a neural network $\mathbf{v} = g(\mathcal{A})$, where $\mathcal{A}$ is the set of annotation vectors. For simplicity, we just apply a linear layer, followed by a max-pooling layer to realize the guiding network $g$. Specifically, max-pooling is operated on the elements at the same position of outputs from the linear layer. To introduce more useful external information, we can use the concatenation $\mathbf{b}_i = [\mathbf{a}_i; \mathbf{e}]$ to replace $\mathbf{a}_i$ and form a new set of annotation vectors, where $\mathbf{e}$ is a vector generated by some other process describing the attributes of input image. The output of guiding network is then used to compose the input for decoder. Hence error signal at each time step could be propagated back to the guiding network. Therefore, the guiding network is completed plugged into the current encoder-coder framework and the guiding vector could be learned adaptively.

\subsubsection{Attribute Features}
Generally, the vector $\mathbf{e}$ which introduces external information into the guiding network could be any attribute features extracted from the input image. We can just simply set it as  $\mathbf{a}_0$. However, the more informative $\mathbf{e}$ is, the better performance will be achieved. To further improve the quality of generated captions, we adopt the predictions of frequent word occurrences in captions, which has been proved to be powerful \cite{jia_iccv2015,fang2015captions}. The word occurrence prediction could be obtained by the weakly-supervised approach of Multiple Instance Learning (MIL) \cite{viola2005multiple}. In this paper, we adopt the same model as in \cite{fang2015captions}. We denote the word occurrence predictions as MIL in this paper.

\subsubsection{Discriminative Training}
Discriminative supervision has been proved to be useful in \cite{fang2015captions}. Our guiding network provides a natural way to incorporate discriminative supervision. We can just treat the elements of guiding vector as scores for words and force the words occurred in the final caption to have higher scores. We adopt the margin-based multiple label criterion and the loss function of discriminative supervision for one $<$\texttt{image,caption}$>$ pair could be expressed as
\begin{align}
	\mathcal{L}_{dis} =\frac{1}{Z} \sum_{j \in \mathcal{C}} \sum_{i \notin \mathcal{C}} \max(0, 1-(s_j - s_i)),
\end{align}
where $\mathcal{C}$ is the set of all frequent words in the current caption, $Z$ is the dimensionality of guiding vector and $s_i$ is the $i$-th element of guiding vector. Combining the negative log-likelihood in Eq.~\eqref{obj_lm}, we can obtain the objective function for one  $<$\texttt{image,caption}$>$ pair:
\begin{align}\label{obj_lstm}
	\mathcal{L}_{all} = \mathcal{L}  + \lambda \mathcal{L}_{dis},  
\end{align}
where $\lambda$ is a trade-off parameter between the language model and the discriminative supervision. To reduce the non-expected effects from less common words, we use top 1,000 frequent words in the captions of training set in the discriminative training process.

\subsection{Extension to Review Net} 
Review Net \cite{yang2016review} is an enhanced extension of the encoder-decoder framework. It performs a number of review steps with attention mechanism on the annotation vectors from encoder, and outputs a thought vector at each review step; the thought vectors are then used as the input of the attention model in the decoder step. The review step can be seen as a process of learning the initial states and annotation vectors for decoder. But the Review Net does not learn the guiding information. It is the current state-of-the-art model for image captioning. 

\begin{figure}[h] 
 \centering
  \includegraphics[width=0.9\linewidth]{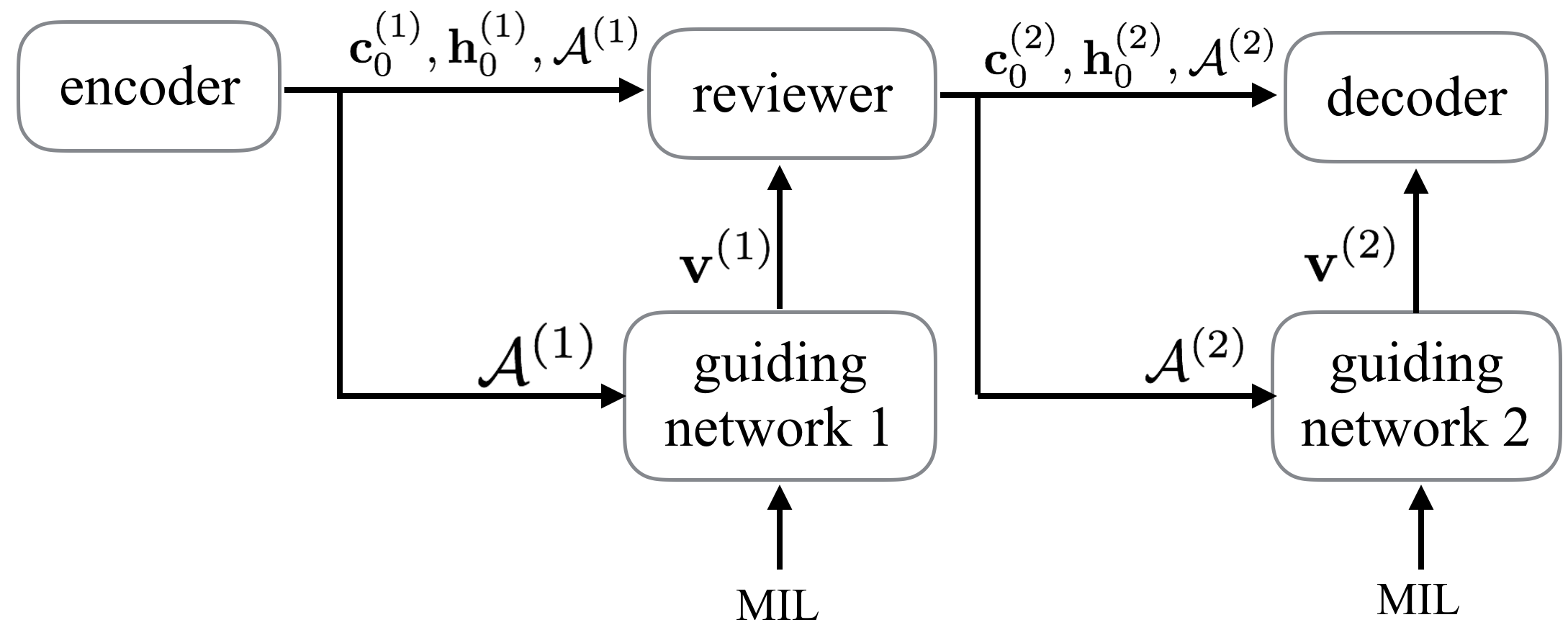}
  \caption{Learning to guide for Review Net. Two guiding networks are employed to generate guiding vectors for review step and decoder step.}
  \label{lg-review-net}
\end{figure}

The review step in Review Net is a special kind of decoder. The input vector $\mathbf{x}_t$ at each review step is just zero vector and the parameters of LSTM unit are not shared\footnote{Generally, for the review step, the parameters of LSTM can either be shared among all the units or not. Usually, Review Net with shared parameters at review steps performs better.}. We can extend our idea to the Review Net easily, and the main framework is shown in Figure~\ref{lg-review-net}. In this extension,  we employ two guiding networks. One is for the review step and its inputs are MIL and the annotation vectors $\mathcal{A}^{(1)}$ from the encoder steps. Its output,  the guiding vector $\mathbf{v}^{(1)}$, is used as input for each review step. Another guiding network is for the decoder step. The newly generated annotation vectors $\mathcal{A}^{(2)}$ and MIL are used as inputs, and its output $\mathbf{v}^{(2)}$ is used to guide the decoder. We also use the discriminative supervision for both guiding networks in Review Net. The complete objective function for one $<$\texttt{image,caption}$>$  pair could be written as:
\begin{align}\label{obj_review_net}
	\mathcal{L}_{all} = \mathcal{L}  + \lambda_1 \mathcal{L}_{dis1} + \lambda_2 \mathcal{L}_{dis2},
\end{align}
where $\mathcal{L}_{dis1}$  and $\mathcal{L}_{dis2}$ are discriminative loss functions for the guiding networks in review step and decoder step respectively, and $\lambda_1$ and $\lambda_2$ are trade-off parameters. In practical applications, we just set them with the same value.


\section{Experiments}\label{experiment}

\begin{table*}[t]
\centering
 \small
\caption{Single model performance of different image captioning models on the MS COCO dataset. The highest value of each entry has been highlighted in boldface.}
\label{single_result}
\begin{tabular}{l|c|c|c|c|c|c|c}
\hline
\hline
  Image Captioning Model & BLEU-1 & BLEU-2 & BLEU-3 & BLEU-4& METEOR & ROUGE-L & CIDEr\\
  \hline
  \hline
  Soft Attention  \cite{xu2015show} &0.714 &0.544 &0.409&0.308&0.255  &0.530&0.978 \\
  Review Net \cite{yang2016review} &0.729 &0.563&0.427& 0.324 &0.257&0.538 &1.009 \\
  LSTM-A5 \cite{yao2016boosting}&0.722 & 0.550 &0.414&0.312& 0.255 &0.531&0.985\\
  Text Attention \cite{jmun_aaai2017}  &\textbf{0.749} &\textbf{0.581}&0.437&0.326&0.257&-&1.024\\
    Sentence Attention \cite{zhou2016image}&0.716 & 0.545&0.405&0.301&0.247  &-&0.970\\
  Attribute LSTM \cite{wu2016value} &0.740 &0.560&0.420&0.310&0.260&-&0.940\\

RIC \cite{hliu_ric2016}  & 0.734 & 0.535& 0.385&0.299&0.254&-&-\\
RHN \cite{jgu_rhn2016} &0.723 &0.553 &0.413 &0.306&0.252&-&0.989\\
  \hline
  \hline
  MIL-Soft-Attention &0.735&0.570&0.432&0.327&0.257&0.540&1.005\\
  MIL-Review-Net &0.733&0.568&0.431&0.327&0.258&0.542&1.012\\
  \hline
  \hline
  LTG-Soft-Attention  &0.732&0.565&0.428&0.323&0.259&0.537&1.023\\
  LTG-Review-Net  &0.743&0.579&\textbf{0.442}&\textbf{0.336}&\textbf{0.261}&\textbf{0.548}&\textbf{1.039}\\
\hline
\hline
\end{tabular}
\end{table*}

\begin{table*}[h]
\centering
 \small
\caption{Performance of ensemble models on the MS COCO dataset. The highest value of each entry has been highlighted.}
\label{ensemble_result}
\begin{tabular}{l|c|c|c|c|c|c|c}
\hline
\hline
  Image Captioning Model & BLEU-1 & BLEU-2 & BLEU-3 & BLEU-4& METEOR & ROUGE-L & CIDEr\\
  \hline
  \hline
  Soft Attention  \cite{xu2015show} &0.731 &0.564  &0.430&0.328&0.257  &0.540&1.015\\
  Review Net \cite{yang2016review} &0.738 &0.575 &0.440&0.336& 0.262 &0.547& 1.046\\
  LSTM-A5 \cite{yao2016boosting}&0.730 &0.565  &0.429 &0.325 &0.251  &0.538 &0.986\\
  Attribute Attention \cite{you2016image}  &0.709 &0.537 &0.402&0.304 & 0.243  &-&-\\
  NIC \cite{vinyals2015show}  & - & - & -&0.321&0.257  &-&0.998\\
  \hline
  \hline
 MIL-Soft-Attention &0.745&0.583&0.447&0.342&0.264& 0.550 &1.058\\
 MIL-Review-Net  &0.747&0.584&0.449&0.344&0.264& 0.551 &1.063\\
  \hline
  \hline
LTG-Soft-Attention &0.747&0.586&0.451&0.347&0.265&0.552  &1.070\\
LTG-Review-Net &\textbf{0.751}&\textbf{0.592}&\textbf{0.458}&\textbf{0.353}&\textbf{0.267}&\textbf{0.555}  &\textbf{1.078}\\
\hline
\hline
\end{tabular}
\end{table*}

In this section, we first introduce the dataset and experiment settings. Afterwards, we demonstrate the effectiveness of our proposed model, and present comparisons with the state-of-the-art image captioning models.

\subsection{Dataset}
The MS COCO dataset\footnote{http://mscoco.org/} \cite{lin2014microsoft} is the most popular benchmark dataset for image captioning, which contains about 123,000 images, with each image associated with five captions. Following the conventional evaluation procedure \cite{jmun_aaai2017,yao2016boosting,yang2016review}, we employ the same data split as in \cite{karpathy2015deep} for the performance comparisons, where  5,000 images as well as the associated captions are reserved as validation and test set, respectively, with the rest employed for training. MIL model are trained on the original training split which contains about 82,000 images.

For the captions, we discard all the non-alphabetic characters, transform all letters into lowercase, and tokenize the captions using white space. Moreover, all the words with the occurrences less than 5 times are replaced by the unknown token $\texttt{<UNK>}$. Thus a vocabulary consisting of 9,520 words is finally constructed. Furthermore, we truncate all the captions longer than 30 tokens.

\subsection{Configurations and Settings}

Inception-V3 \cite{szegedy2016rethinking} is used as encoder to extract image  representation with  dimension  2048  and annotation vectors with  dimension  $64\times 1280$.  The LSTM size is set as 2048. The parameters of LSTM are initialized with uniform distribution in $[-0.1, 0.1]$. The AdaGrad \cite{duchi2011adaptive} is applied to optimize the network, and learning rate is set as 0.01 and weight decay is set as $10^{-4}$. Early stopping strategy is used to prevent overfitting. If the evaluation measurement on validation set, specifically the CIDEr, reaches the maximum value, we terminate the training procedure and use the corresponding model for further testing.

For sentence generation in testing stage, there are two common strategies. The first one is greedy search, which  choose the word with maximum probability at each time step and set it as LSTM input for next time step until the end-of-sentence sign is emitted or the maximum length of sentence is reached. The second one is the beam search strategy which selects the top-$k$ best sentences at each time step and considers them as the candidates to generate new top-$k$ best sentences at the next time step. Usually beam search strategy provides better performance \cite{vinyals2016show}, hence we adopt this strategy in our experiments. And the $k$ for all experiments are set as 3.

\subsection{Experimental Results and Analysis}

We compare our proposed model with state-of-the-art approaches on image captioning, including Neural Image Caption (NIC)~\cite{vinyals2015show},  Attribute LSTM~\cite{wu2016value}, LSTM-A5~\cite{yao2016boosting}, Recurrent Image Captioner (RIC)~\cite{hliu_ric2016}, Recurrent Highway Network (RHN)~\cite{jgu_rhn2016}, Soft Attention model~\cite{xu2015show}, Attribute Attention model~\cite{you2016image}, Sentence Attention model~\cite{zhou2016image}, Review Net~\cite{yang2016review}, and Text Attention model~\cite{jmun_aaai2017}. 

Following the standard evaluation process, four types of metrics are used for performance comparisons, specifically the BLEU \cite{papineni2002bleu},  METEOR \cite{banerjee2005meteor}, ROUGE-L~\cite{lin2004rouge}, and CIDEr \cite{vedantam2015cider}. These metrics measure the n-gram occurrences in the generated sentence from each image captioning model and the ground truth sentence, with the consideration of the n-gram saliency and rarity. We use the official MS COCO caption evaluation scripts\footnote{https://github.com/tylin/coco-caption} for the performance comparisons. 

We report the performance of our method combined with Soft Attention \cite{xu2015show} and Review Net \cite{yang2016review}, which are denoted as LTG-Soft-Attention and LTG-Review-Net, respectively. To prove that the learned guiding vector is better than MIL, we combine MIL with those two methods which are denoted as MIL-Soft-Attention and MIL-Review-Net, respectively. For Soft Attention, the guiding vector is replaced with MIL in Eq.~\eqref{input_of_decoder}. For Review Net, besides the above operation for the decoder steps, we also replace guiding vector with MIL in the input for the LSTM in the review steps. The settings, e.g. data split, type of encoder or metrics used, of Soft Attention \cite{xu2015show}, Review Net \cite{yang2016review} and LSTM-A5 \cite{yao2016boosting} are different from ours. For fair comparison, \textit{we reimplemented these methods and the performance under the same settings are provided}. For other methods, we report the published results. Note that Attribute LSTM~\cite{wu2016value}, RCI~\cite{hliu_ric2016} and RHN~\cite{jgu_rhn2016} used VGG net, while Text Attention~\cite{jmun_aaai2017} used GoogLeNet. The performance of different models are provided in Table~\ref{single_result}. We can see that methods with guiding information (MIL of LTG) provide better performance than their counterpart without such information, which proves that injecting useful information into the decoder helps improve the quality of captions. Moreover, our LTG-Soft-Attention and LTG-Review-Net perform better than MIL-Soft-Attention and MIL-Review-Net respectively, which shows that the learned guiding vector is more effective. LTG-Review-Net outperforms the other models on all metrics except BLEU-1 and BLEU-2 of Text Attention. Text Attention uses much more powerful ResNet \cite{he2016deep} as encoder, which helps to boost captioning performance. Moreover, sentences retrieved based on image similarity are used in the attention model. It seems that such attention model favors BLEU-1 and BLEU-2.

Following the common strategy \cite{vinyals2015show}, ensembles of five models trained with different initializations are further employed to examine the sentence generation ability. The performance comparisons are illustrated in Table~\ref{ensemble_result}. For our reimplementations of Soft Attention and Review Net, we employ Inception-V3 as the image encoder, while NIC, the Attribute Attention model, and LSMT-A5 use GoogleNet, and the Sentence Attention model uses VGG. Note that the results of NIC are obtained on a different testing set. It can be observed that the results of LTG-Review-Net outperforms all the other competitor approaches on all measurements.  This superiority mainly attributes to the adaptively learned guiding vector.

\begin{figure*}[t]
 \centering
  \begin{tabular}{ | c | m{6.8cm} | m{6.8cm} | }
    \hline
    Image & Captions generated & Ground truth \\
    \hline
    \hline
    
    \begin{minipage}{.15\textwidth}
      \includegraphics[width=\linewidth, height=0.8\linewidth]{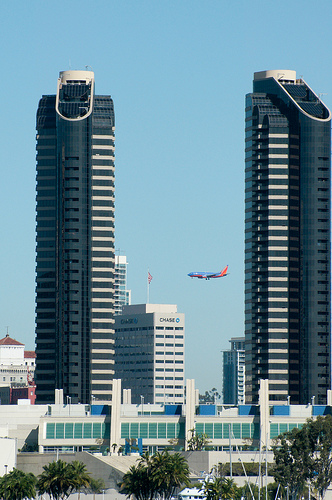}
    \end{minipage}
    &
	\begin{minipage}[t]{6.8cm}\scriptsize
        \textbf{LTG-Review-Net}: a {\color{red}plane} flying over a city with tall buildings. \\
        \textbf{MIL-Review-Net}: a bunch of {\color{blue}kites} flying over a city. \\
        \textbf{MIL-Soft-Attention}: a group of {\color{blue}kites} flying in the sky. \\
        \textbf{Review Net}: a group of {\color{blue}people} flying {\color{blue}kites} in the air. \\
        \textbf{Soft attention}: a group of {\color{blue}people} flying {\color{blue}kites} in the air.  \\
        \textbf{LSTM-A5}: a {\color{red}plane} flying over a city with tall buildings.	
    \end{minipage}
    &
    \begin{minipage}{6.8cm}\scriptsize
1. a distant airplane flying between two large buildings. \\
2. a couple of tall buildings with a large jetliner between them. \\
3. commercial airliner flying between two tall structures on clear day. \\
4. two tall buildings and a plane in between them. \\
5. an airplane is seen between two identical skyscrapers.
    \end{minipage}
    \\
    \hline
    \hline 
    
    \begin{minipage}{.15\textwidth}
      \includegraphics[width=\linewidth, height=0.8\linewidth]{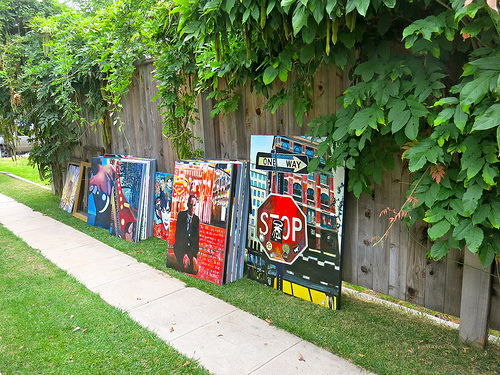}
    \end{minipage}
    &
	\begin{minipage}[t]{6.8cm}\scriptsize
        \textbf{LTG-Review-Net}: a bunch of signs that are sitting in the grass. \\
        \textbf{MIL-Review-Net}: a sign that is on the side of a building.\\
        \textbf{MIL-Soft-Attention}: a {\color{blue}fire hydrant} on the side of the road.\\
        \textbf{Review Net}: a group of {\color{blue}people} standing next to a red {\color{blue}fire hydrant}. \\
        \textbf{Soft Attention}: a street scene with a {\color{blue}fire truck and a fire hydrant}. \\
        \textbf{LSTM-A5}: a sign that is on the side of a road.	
    \end{minipage}
    &
    \begin{minipage}{6.8cm}\scriptsize
1. a collection of artwork leaning against a wooden fence. \\
2. a collection of poster arts lined up on the fence. \\
3. a collection of paintings against a fence outside. \\
4. several paintings leaning against a fence in the grass. \\ 
5. large set of personal paintings sitting on the side of a fence.
    \end{minipage}
    \\
    \hline
    \hline      
    
    \begin{minipage}{.15\textwidth}
      \includegraphics[width=\linewidth, height=0.8\linewidth]{COCO_val2014_000000484301.jpg}
    \end{minipage}
    &
	\begin{minipage}[t]{6.8cm}\scriptsize
        \textbf{LTG-Review-Net}: a {\color{red}parking lot} filled with {\color{red}cars} and {\color{red}motorcycles}.\\
        \textbf{MIL-Review-Net}: a group of {\color{red}motorcycles} parked next to each other.\\
        \textbf{MIL-Soft-Attention}: a group of {\color{red}motorcycles} parked in a {\color{red}parking lot}.\\
        \textbf{Review Net}: a {\color{blue}man riding} a motorcycle down a street.\\
        \textbf{Soft Attention}: a {\color{blue}man riding} a motorcycle down a street.	 \\
        \textbf{LSTM-A5}: a {\color{red}motorcycle} is parked on the side of the street.  	
    \end{minipage}
    &
    \begin{minipage}{6.8cm}\scriptsize
1. a couple of motor bikes and cars on a street. \\
2. a pair of motorcycles parked in a bike slot. \\
3. a busy day on the street including cars and motorcycles. \\
4. motorcycles parked in designated spaces in a parking lot. \\
5. a parking lot scene with cars and motorcycles.
    \end{minipage}
    \\
    \hline
    \hline 
    
    \begin{minipage}{.15\textwidth}
      \includegraphics[width=\linewidth, height=0.8\linewidth]{545007.jpg}
    \end{minipage}
    &
	\begin{minipage}[t]{6.8cm}\scriptsize
        \textbf{LTG-Review-Net}: a {\color{red}statue} of a clock on top of a tower. \\
        \textbf{MIL-Review-Net}: a clock tower with a clock on it.\\
        \textbf{MIL-Soft-Attention}: a clock tower with a statue on top of it.\\
        \textbf{Review Net}: a black and white clock on a tower. \\
        \textbf{Soft Attention}: a black and white photo of a clock on top of a building. \\
        \textbf{LSTM-A5}: a clock tower with a {\color{red}statue} of a man on top of it.  	
    \end{minipage}
    &
    \begin{minipage}{6.8cm}\scriptsize
1. a clock tower with a statue on top of it. \\
2. a statue holding a glass light stands atop an old clock. \\
3. a clock with a statue of a woman on top. \\
4. a statue is standing on top of a clock on a pole. \\
5. a statue holding a lamp on top of a clock. 
    \end{minipage}
    \\
    \hline
    \hline 
        
    \begin{minipage}{.15\textwidth}
      \includegraphics[width=\linewidth, height=0.8\linewidth]{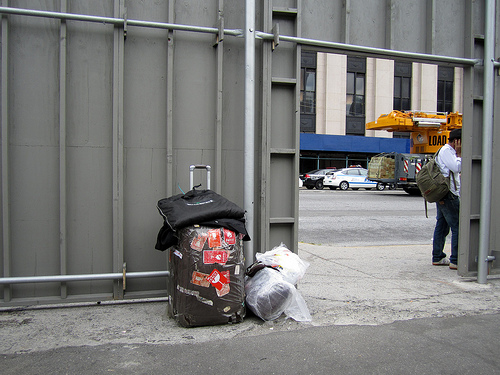}
    \end{minipage}
    &
	\begin{minipage}[t]{6.8cm}\scriptsize
        \textbf{LTG-Review-Net}: a couple of {\color{red}bags} that are on the ground. \\
        \textbf{MIL-Review-Net}: a person standing next to a pile of {\color{red}luggage}.\\
        \textbf{MIL-Soft-Attention}: a {\color{blue}trash can} a trash can and a trash can.\\
        \textbf{Review Net}: a {\color{blue}fire hydrant} sitting on the side of a road.\\
        \textbf{Soft Attention}: a {\color{blue}fire hydrant} sitting on the side of a street. \\
        \textbf{LSTM-A5}: a street scene with a {\color{blue}trash can} and a garbage can.	
    \end{minipage}
    &
    \begin{minipage}{6.8cm}\scriptsize
1. a couple of bags of someone's belongings that were left unattended. \\
2. luggage hidden behind a fence while a man makes a cell phone call \\
3. a piece of luggage sitting up against a gray wall.\\
4. a person with a backpack is standing next to a doorway with luggage on the other side. \\
5. a man on the phone near luggage by a gate.
    \end{minipage}
    \\
    \hline
    \hline 

  \end{tabular}
  \caption{Captions generated by LTG-Review-Net, MIL-Review-Net, MIL-Soft-Attention, Review Net, Soft Attention and LSTM-A5, as well as the corresponding ground truths. The elements that are correctly recognized and mentioned in text are marked in red, while the elements wrongly recognized are marked in blue.}
  \label{caption-example}
\end{figure*}

\paragraph{Qualitative analysis.}
The qualitative results generated by the single model of LTG-Review-Net, MIL-Review-Net, MIL-Soft-Attention, Review Net, Soft Attention and LSTM-A5 are illustrated Table~\ref{caption-example}. We select some typical images on which these methods behave differently and present the captions generated and corresponding ground truth. We can see that Review Net and Soft Attention do not generate the correct captions for the first three images. The salient objects recognized by them do not even exist in the images. There is no ``people'' in the first image, no ``hydrant'' in the second image and no ``man'' in the third image. Besides, with the help of MIL, the ability of recognizing objects is improved. For example, Review Net and Soft Attention generate ``people'', while their counterparts with MIL do not. Even with the help of MIL, Review Net failed to recognize ``plane'' in the first image and ``car'' in the third image.  For the fourth image, Review Net and Soft Attention fail to recognize ``statue''.  For the fifth image, 
 Review Net, MIL-Soft-Attention and Soft Attention also fail to provide satisfactory results. Moreover, LSTM-A5 also performs badly.  They recognize the wrong objects for the fifth image, like Review Net and Soft Attention do.  
 MIL-Review-Net recognizes the ``luggage'' in the fifth image,
 which should be attributed to the help from MIL. 
From the above analysis, we can see that the learned guiding vector could help the model to recognize the correct salient objects. We believe that that is one of the reasons why guiding network could help to improve the quality of descriptions.

 \begin{table}[h]
\centering
\caption{Performance with different weights for discriminative supervision.}
\label{weight_effect}
\begin{tabular}{l|cc}
  \hline   \hline
 Weights & CIDEr & BLEU-4 \\
  \hline   \hline
 100 &0.963 & 0.306\\
 10 & 0.983 &0.309\\
 1 &0.964 &  0.303\\
 0.1 &0.947 & 0.300\\
 0.01 &0.939  & 0.293\\
   \hline \hline
\end{tabular}
\end{table}

\paragraph{The effect of discriminative supervision.}
We study the effect of the discriminative supervision in this subsection. We set the same value for $\lambda_1$ and $\lambda_2$ in Eq.~\eqref{obj_review_net} and train the LTG-Review-Net with different  trade-off parameters. The performance\footnote{The greedy search was used and the average performance of 8 models with different initialization was reported.} on validation set are presented in Table~\ref{weight_effect}.  We can see that the discriminative training with an appropriate weight do help improve performance. Our method is not very sensitive to the weight of discriminative loss and $10$ was used in all our experiments.

\begin{table}[h]
\centering
\caption{Performance of our methods with or without MIL and annotation vectors. }
\label{contributions_parts}
\begin{tabular}{l|lcc}
  \hline  \hline
Models & Settings & CIDEr & BLEU-4 \\
  \hline
    \hline
LTG-Review-Net &Keep MIL &0.975& 0.304\\
LTG-Review-Net &Keep $\mathcal{A}$ &0.973 &0.307 \\
LTG-Review-Net &Keep both &0.983 &0.309 \\
LTG-Review-Net &Keep None &0.958 &  0.302\\  
  \hline
LTG-Soft-Attention &Keep MIL&0.931&0.292 \\
LTG-Soft-Attention &Keep $\mathcal{A}$&	0.936& 0.291\\
LTG-Soft-Attention &Keep both &0.965 &0.301 \\
LTG-Soft-Attention &Keep None &0.929 &0.287 \\
  \hline  \hline
\end{tabular}
\end{table}

\begin{table}[h]
\centering
\caption{Number of distinct words in the captions from different models.}
\label{word_num}
\begin{tabular}{l|c}  
\hline
  \hline
 Model & Number of Words \\
  \hline
    \hline
LTG-Review-Net & 840\\
LTG-Soft-Attention & 938\\
MIL-Review-Net & 745\\
MIL-Soft-Attention &782 \\
Review Net &762 \\
Soft Attention &793 \\
LSTM-A5 & 826\\
  \hline
    \hline
\end{tabular}
\end{table}

\paragraph{Contributions of MIL and Annotation Vectors.}
We set either MIL or annotation vectors to be zero vector to study their contributions for the performance. For LTG-Review-Net, we only consider the annotation vectors generated by encoder and the learned annotation vectors by review steps are not considered in this experiment. The performance of different models under different settings is listed in Table~\ref{contributions_parts}. We can see that it is difficult to tell which one is more important since the performance of models with either MIL or annotation vectors is close. Moreover, the guiding network without MIL and annotations performed the worst and guiding network with both performed the best. It is quite natural since more input information usually leads to better performance. If more informative features are provided, we can just concat them with the current inputs of the guiding network to form new inputs.

\paragraph{Richness of Words.}
We computed the number of distinct words in the captions generated by different models on the testing set and the results are presented in Table~\ref{word_num}. We can see that the combinations of guiding network with Review Net and Soft Attention generate much more words than the other models. This phenomenon shows that guiding network might help decoder to select more words with higher accuracies. This should be attributed to that the learned guiding vector contains language information.


\section{Conclusion}\label{conclusion}
In this work, we proposed to learn a guiding vector with a guiding neural network for the decoder step. The learned guiding vector is used to compose the input of the decoder. This guiding network can be plugged into the current encode-decoder framework and the guiding vector can be learned adaptively, making itself to embed information of both image and language. Thus, more information is injected into the decoder so that balancing the factors of describing the input image and fitting the language model during the caption generation process becomes easier, which help improve the quality of captions generated. The experiments performed on the MS COCO dataset verify the advantages of our approach.
{\small
\bibliographystyle{aaai}
\bibliography{ltg_ref}

\begin{thebibliography}{}

\bibitem[\protect\citeauthoryear{Banerjee and Lavie}{2005}]{banerjee2005meteor}
Banerjee, S., and Lavie, A.
\newblock 2005.
\newblock Meteor: An automatic metric for mt evaluation with improved
  correlation with human judgments.
\newblock In {\em ACL workshop}.

\bibitem[\protect\citeauthoryear{Chen \bgroup et al\mbox.\egroup
  }{2017}]{lchen_attention2016}
Chen, L.; Zhang, H.; Xiao, J.; Nie, L.; Shao, J.; Liu, W.; and Chua, T.-S.
\newblock 2017.
\newblock Sca-cnn: Spatial and channel-wise attention in convolutional networks
  for image captioning.
\newblock In {\em CVPR}.

\bibitem[\protect\citeauthoryear{Chung \bgroup et al\mbox.\egroup
  }{2014}]{chung2014empirical}
Chung, J.; Gulcehre, C.; Cho, K.; and Bengio, Y.
\newblock 2014.
\newblock Empirical evaluation of gated recurrent neural networks on sequence
  modeling.
\newblock {\em arXiv preprint arXiv:1412.3555}.

\bibitem[\protect\citeauthoryear{Duchi, Hazan, and
  Singer}{2011}]{duchi2011adaptive}
Duchi, J.; Hazan, E.; and Singer, Y.
\newblock 2011.
\newblock Adaptive subgradient methods for online learning and stochastic
  optimization.
\newblock {\em Journal of Machine Learning Research} 12(Jul):2121--2159.

\bibitem[\protect\citeauthoryear{Fang \bgroup et al\mbox.\egroup
  }{2015}]{fang2015captions}
Fang, H.; Gupta, S.; Iandola, F.; Srivastava, R.~K.; Deng, L.; Doll{\'a}r, P.;
  Gao, J.; He, X.; Mitchell, M.; Platt, J.~C.; et~al.
\newblock 2015.
\newblock From captions to visual concepts and back.
\newblock In {\em CVPR},  1473--1482.

\bibitem[\protect\citeauthoryear{Farhadi \bgroup et al\mbox.\egroup
  }{2010}]{farhadi2010every}
Farhadi, A.; Hejrati, M.; Sadeghi, M.~A.; Young, P.; Rashtchian, C.;
  Hockenmaier, J.; and Forsyth, D.
\newblock 2010.
\newblock Every picture tells a story: Generating sentences from images.
\newblock In {\em ECCV}.

\bibitem[\protect\citeauthoryear{Gu, Wang, and Chen}{2016}]{jgu_rhn2016}
Gu, J.; Wang, G.; and Chen, T.
\newblock 2016.
\newblock Recurrent highway networks with language cnn for image captioning.
\newblock {\em arXiv:1612.07086}.

\bibitem[\protect\citeauthoryear{He \bgroup et al\mbox.\egroup
  }{2016}]{he2016deep}
He, K.; Zhang, X.; Ren, S.; and Sun, J.
\newblock 2016.
\newblock Deep residual learning for image recognition.
\newblock In {\em CVPR},  770--778.

\bibitem[\protect\citeauthoryear{Hochreiter and
  Schmidhuber}{1997}]{hochreiter1997long}
Hochreiter, S., and Schmidhuber, J.
\newblock 1997.
\newblock Long short-term memory.
\newblock {\em Neural computation} 9(8):1735--1780.

\bibitem[\protect\citeauthoryear{Jia \bgroup et al\mbox.\egroup
  }{2015}]{jia_iccv2015}
Jia, X.; Gavves, E.; Fernando, B.; and Tuytelaars, T.
\newblock 2015.
\newblock Guiding long-short term memory for image caption generation.
\newblock {\em ICCV}.

\bibitem[\protect\citeauthoryear{Karpathy and Fei-Fei}{2015}]{karpathy2015deep}
Karpathy, A., and Fei-Fei, L.
\newblock 2015.
\newblock Deep visual-semantic alignments for generating image descriptions.
\newblock In {\em CVPR},  3128--3137.

\bibitem[\protect\citeauthoryear{Kulkarni \bgroup et al\mbox.\egroup
  }{2013}]{kulkarni2013babytalk}
Kulkarni, G.; Premraj, V.; Ordonez, V.; Dhar, S.; Li, S.; Choi, Y.; Berg,
  A.~C.; and Berg, T.~L.
\newblock 2013.
\newblock Babytalk: Understanding and generating simple image descriptions.
\newblock {\em IEEE Transactions on Pattern Analysis and Machine Intelligence}
  35(12):2891--2903.

\bibitem[\protect\citeauthoryear{Kuznetsova \bgroup et al\mbox.\egroup
  }{2012}]{kuznetsova2012collective}
Kuznetsova, P.; Ordonez, V.; Berg, A.~C.; Berg, T.~L.; and Choi, Y.
\newblock 2012.
\newblock Collective generation of natural image descriptions.
\newblock In {\em ACL}.

\bibitem[\protect\citeauthoryear{Lin \bgroup et al\mbox.\egroup
  }{2014}]{lin2014microsoft}
Lin, T.-Y.; Maire, M.; Belongie, S.; Hays, J.; Perona, P.; Ramanan, D.;
  Doll{\'a}r, P.; and Zitnick, C.~L.
\newblock 2014.
\newblock Microsoft coco: Common objects in context.
\newblock In {\em ECCV}.

\bibitem[\protect\citeauthoryear{Lin}{2004}]{lin2004rouge}
Lin, C.-Y.
\newblock 2004.
\newblock Rouge: A package for automatic evaluation of summaries.
\newblock In {\em ACL-04 workshop}.

\bibitem[\protect\citeauthoryear{Liu \bgroup et al\mbox.\egroup
  }{2016}]{hliu_ric2016}
Liu, H.; Yang, Y.; Shen, F.; Duan, L.; and Shen, H.~T.
\newblock 2016.
\newblock Recurrent image captioner: Describing images with spatial-invariant
  trnsformation and attention fieltering.
\newblock {\em arXiv:1612.04949}.

\bibitem[\protect\citeauthoryear{Mason and
  Charniak}{2014}]{mason2014nonparametric}
Mason, R., and Charniak, E.
\newblock 2014.
\newblock Nonparametric method for data-driven image captioning.
\newblock In {\em ACL},  592--598.

\bibitem[\protect\citeauthoryear{Mitchell \bgroup et al\mbox.\egroup
  }{2012}]{mitchell2012midge}
Mitchell, M.; Han, X.; Dodge, J.; Mensch, A.; Goyal, A.; Berg, A.; Yamaguchi,
  K.; Berg, T.; Stratos, K.; and Daum{\'e}~III, H.
\newblock 2012.
\newblock Midge: Generating image descriptions from computer vision detections.
\newblock In {\em ACL}.

\bibitem[\protect\citeauthoryear{Mun, Cho, and Han}{2017}]{jmun_aaai2017}
Mun, J.; Cho, M.; and Han, B.
\newblock 2017.
\newblock Text-guided attention model for image captioning.
\newblock {\em AAAI}.

\bibitem[\protect\citeauthoryear{Papineni \bgroup et al\mbox.\egroup
  }{2002}]{papineni2002bleu}
Papineni, K.; Roukos, S.; Ward, T.; and Zhu, W.-J.
\newblock 2002.
\newblock Bleu: a method for automatic evaluation of machine translation.
\newblock In {\em ACL}.

\bibitem[\protect\citeauthoryear{Rennie \bgroup et al\mbox.\egroup
  }{2017}]{rennie2016self}
Rennie, S.~J.; Marcheret, E.; Mroueh, Y.; Ross, J.; and Goel, V.
\newblock 2017.
\newblock Self-critical sequence training for image captioning.
\newblock In {\em The IEEE Conference on Computer Vision and Pattern
  Recognition (CVPR)}.

\bibitem[\protect\citeauthoryear{Simonyan and Zisserman}{2015}]{Simonyan14c}
Simonyan, K., and Zisserman, A.
\newblock 2015.
\newblock Very deep convolutional networks for large-scale image recognition.
\newblock {\em ICLR}.

\bibitem[\protect\citeauthoryear{Szegedy \bgroup et al\mbox.\egroup
  }{2016}]{szegedy2016rethinking}
Szegedy, C.; Vanhoucke, V.; Ioffe, S.; Shlens, J.; and Wojna, Z.
\newblock 2016.
\newblock Rethinking the inception architecture for computer vision.
\newblock In {\em CVPR}.

\bibitem[\protect\citeauthoryear{Vedantam, Lawrence~Zitnick, and
  Parikh}{2015}]{vedantam2015cider}
Vedantam, R.; Lawrence~Zitnick, C.; and Parikh, D.
\newblock 2015.
\newblock Cider: Consensus-based image description evaluation.
\newblock In {\em CVPR}.

\bibitem[\protect\citeauthoryear{Vinyals \bgroup et al\mbox.\egroup
  }{2015}]{vinyals2015show}
Vinyals, O.; Toshev, A.; Bengio, S.; and Erhan, D.
\newblock 2015.
\newblock Show and tell: A neural image caption generator.
\newblock In {\em CVPR},  3156--3164.

\bibitem[\protect\citeauthoryear{Vinyals \bgroup et al\mbox.\egroup
  }{2017}]{vinyals2016show}
Vinyals, O.; Toshev, A.; Bengio, S.; and Erhan, D.
\newblock 2017.
\newblock Show and tell: Lessons learned from the 2015 mscoco image captioning
  challenge.
\newblock {\em IEEE transactions on pattern analysis and machine intelligence}
  39(4):652--663.

\bibitem[\protect\citeauthoryear{Viola \bgroup et al\mbox.\egroup
  }{2005}]{viola2005multiple}
Viola, P.; Platt, J.~C.; Zhang, C.; et~al.
\newblock 2005.
\newblock Multiple instance boosting for object detection.
\newblock In {\em NIPS}, volume~2, ~5.

\bibitem[\protect\citeauthoryear{Williams}{1992}]{williams1992simple}
Williams, R.~J.
\newblock 1992.
\newblock Simple statistical gradient-following algorithms for connectionist
  reinforcement learning.
\newblock {\em Machine learning} 8(3-4):229--256.

\bibitem[\protect\citeauthoryear{Wu \bgroup et al\mbox.\egroup
  }{2016}]{wu2016value}
Wu, Q.; Shen, C.; Liu, L.; Dick, A.; and van~den Hengel, A.
\newblock 2016.
\newblock What value do explicit high level concepts have in vision to language
  problems?
\newblock In {\em CVPR}.

\bibitem[\protect\citeauthoryear{Xu \bgroup et al\mbox.\egroup
  }{2015}]{xu2015show}
Xu, K.; Ba, J.; Kiros, R.; Cho, K.; Courville, A.~C.; Salakhutdinov, R.; Zemel,
  R.~S.; and Bengio, Y.
\newblock 2015.
\newblock Show, attend and tell: Neural image caption generation with visual
  attention.
\newblock In {\em ICML}.

\bibitem[\protect\citeauthoryear{Yang \bgroup et al\mbox.\egroup
  }{2016}]{yang2016review}
Yang, Z.; Yuan, Y.; Wu, Y.; Cohen, W.~W.; and Salakhutdinov, R.~R.
\newblock 2016.
\newblock Review networks for caption generation.
\newblock In {\em NIPS}.

\bibitem[\protect\citeauthoryear{Yao \bgroup et al\mbox.\egroup
  }{2016}]{yao2016boosting}
Yao, T.; Pan, Y.; Li, Y.; Qiu, Z.; and Mei, T.
\newblock 2016.
\newblock Boosting image captioning with attributes.
\newblock {\em arXiv:1611.01646}.

\bibitem[\protect\citeauthoryear{You \bgroup et al\mbox.\egroup
  }{2016}]{you2016image}
You, Q.; Jin, H.; Wang, Z.; Fang, C.; and Luo, J.
\newblock 2016.
\newblock Image captioning with semantic attention.
\newblock In {\em CVPR},  4651--4659.

\bibitem[\protect\citeauthoryear{Zhou \bgroup et al\mbox.\egroup
  }{2016}]{zhou2016image}
Zhou, L.; Xu, C.; Koch, P.; and Corso, J.~J.
\newblock 2016.
\newblock Watch what you just said: Image captioning with text-conditional
  attention.
\newblock {\em arXiv:1606.04621}.

\end{thebibliography}
}
\end{document}